\pdfoutput=1
\documentclass[11pt]{article}

\usepackage[]{emnlp2022}
\usepackage{longtable, booktabs}
\usepackage{times}
\usepackage{latexsym}
\usepackage{fixltx2e}
\usepackage{amsmath, amsthm, amssymb}
\usepackage{xcolor}

\usepackage{tikz}
\usetikzlibrary{arrows,shapes,snakes,er,backgrounds}
\tikzstyle{block}=[draw opacity=0.7,line width=1.4cm]

\usepackage{array,arydshln,xcolor}

\newcommand\Fontvim{\fontsize{7}{5.2}\selectfont}

\definecolor{mymagenta}{RGB}{170,60,176}
\definecolor{newblue}{RGB}{175,205,254}
\definecolor{newerblue}{RGB}{182,195,247}

\definecolor{myred}{RGB}{145,0,23}

\definecolor{mynewred}{RGB}{188,34,48}
\definecolor{mynewred2}{RGB}{98,0,18}
\definecolor{mynewred3}{RGB}{143,0,26}
\definecolor{mynewblue}{RGB}{108,158,250}
\definecolor{mynewblue2}{RGB}{12,53,163}

\definecolor{transformerff}{RGB}{184,228,245}
\definecolor{transformersa}{RGB}{254,220,177}
\definecolor{transformeremb}{RGB}{250,217,219}
\definecolor{transformeradnorm}{RGB}{240,243,183}
\definecolor{transformerlinear}{RGB}{212,216,234}
\definecolor{transformersoftmax}{RGB}{195,227,198}

\definecolor{encoder}{RGB}{228,242,223}
\definecolor{decoder}{RGB}{250,226,239}

\definecolor{input}{RGB}{163,216,141}
\definecolor{saoutput}{RGB}{241,173,207}
\definecolor{ffoutput}{RGB}{116,178,250}

\definecolor{mynewgray}{RGB}{108,108,108}

\definecolor{mynewtest}{RGB}{155,108,108}
\definecolor{mymagenta}{RGB}{170,60,176}
\definecolor{newblue}{RGB}{175,205,254}
\definecolor{newerblue}{RGB}{182,195,247}

\definecolor{myred}{RGB}{145,0,23}

\definecolor{mynewred}{RGB}{188,34,48}
\definecolor{mynewred2}{RGB}{98,0,18}
\definecolor{mynewred3}{RGB}{143,0,26}
\definecolor{mynewblue}{RGB}{108,158,250}
\definecolor{mynewblue2}{RGB}{12,53,163}

\definecolor{transformerff}{RGB}{184,228,245}
\definecolor{transformersa}{RGB}{254,220,177}
\definecolor{transformeremb}{RGB}{250,217,219}
\definecolor{transformeradnorm}{RGB}{240,243,183}
\definecolor{transformerlinear}{RGB}{212,216,234}
\definecolor{transformersoftmax}{RGB}{195,227,198}

\definecolor{encoder}{RGB}{228,242,223}
\definecolor{decoder}{RGB}{250,226,239}

\definecolor{input}{RGB}{163,216,141}
\definecolor{saoutput}{RGB}{241,173,207}
\definecolor{ffoutput}{RGB}{116,178,250}

\definecolor{mynewgray}{RGB}{108,108,108}

\usepackage[T1]{fontenc}
\usepackage[utf8]{inputenc}
\usepackage{microtype}

\usepackage{tabularx}
\usepackage[titletoc]{appendix}
\usepackage{graphicx}
\usepackage[bottom]{footmisc}

\usepackage{pgfplots}
\usetikzlibrary{positioning}

\graphicspath{ {./} }
\newcommand\BibTeX{B\textsc{ib}\TeX}

\title{Hierarchical Multi-Label Classification of Scientific Documents}

\author{Mobashir Sadat \mbox{   }\mbox{   }\mbox{   }\mbox{   } Cornelia Caragea\\
  Computer Science \\
  University of Illinois Chicago \\
  \texttt{msadat3@uic.edu \mbox{    }  cornelia@uic.edu} 
  }
\date{}

\begin{document}
\maketitle

\begin{abstract}
Automatic topic classification has been studied extensively to assist managing and indexing scientific documents in a digital collection. With the large number of topics being available in recent years, it has become necessary to arrange them in a hierarchy. Therefore, the automatic classification systems need to be able to classify the documents hierarchically. In addition, each paper is often assigned to more than one relevant topic. For example, a paper can be assigned to several topics in a hierarchy tree. In this paper, we introduce a new dataset for hierarchical multi-label text classification (HMLTC) of scientific papers called \textbf{SciHTC}, which contains $186,160$ papers and $1,233$ categories from the ACM CCS tree. We establish strong baselines for HMLTC and propose a multi-task learning approach for topic classification with keyword labeling as an auxiliary task. Our best model achieves a Macro-F1 score of {$34.57\%$}  which shows that this dataset provides significant research opportunities on hierarchical scientific topic classification. We make our dataset and code available on Github.\footnote{\url{https://github.com/msadat3/SciHTC}}
\end{abstract}

\section{Introduction}
%\vspace{-1.5mm}
With the exponential increase of scientific documents being published every year, the difficulty in managing and categorizing them in a digital collection is also increasing. While the enormity of the number of papers is the most important reason behind it, the problem can also be assigned to the large number of topics. It is a very difficult task to index a paper in a digital collection when there are thousands of topics to choose from. Fortunately, the large number of topics can be arranged in a hierarchy because, except for a few general topics, all topics can be seen as a sub-area of another topic. After arranging the topics in a hierarchy tree, the task of categorizing a paper becomes  much simpler since now there are only a handful of topics to choose from at each level of the hierarchy. However, manual assignment of topics to a large number of papers is still very difficult and expensive, making an automatic system of hierarchical classification of scientific documents a necessity. 

After arranging the topics in a hierarchy, the classification task no longer remains a multi-class classification because in multi-class classification, each paper is classified 
into exactly one of several mutually exclusive classes or topics. However, if the topics are arranged in a hierarchy, they no longer remain mutually exclusive: a paper assigned to a topic node $a$ in the hierarchy also gets assigned to topic node $b$, where $b$ is a node in the parental history of $a$. For example, if a paper is classified to the area of natural language processing (NLP), it is also assigned to the topic of artificial intelligence (AI) given that AI is in the parental history of NLP. This \textit{non-mutual exclusivity} among the topics makes the task a multi-label classification task.

Despite being an important problem, hierarchical multi-label topic classification (HMLTC) has not been explored to a great extent in the context of scientific papers. Most works on hierarchical and/or multi-label topic classification focus on news articles \cite{banerjee-etal-2019-hierarchical, PengDGCNN} and use the RCV-1 dataset \cite{lewis2004rcv1} for evaluation. This is partly because of a lack of datasets for hierarchically classified scientific papers, which hinders progress in this domain. Precisely, the existing multi-label datasets of scientific papers are either comparatively small \cite{kowsari2017HDLTex} or the label hierarchy is not deep \cite{yang-etal-2018-sgm}. 

Therefore, we address the scarcity of datasets for HMLTC on scientific papers by introducing a new large dataset called \textbf{SciHTC} in which the papers are hierarchically classified based on the ACM CCS tree.\footnote{\url{https://dl.acm.org/ccs}} Our dataset is large enough to allow deep learning exploration, comprising $186,160$ research papers that are organized into $1,233$ topics, which are arranged in a six-level deep hierarchy.
We establish several strong baselines for both hierarchical and flat multi-label classification for \textbf{SciHTC}. In addition, we conduct a thorough investigation on the usefulness of author specified keywords in topic classification. Furthermore, we show how multi-task learning with scientific document classification as the principal task and its keyword labeling as the auxiliary task can help improve the classification performance. However, our best models with SciBERT \cite{beltagy-etal-2019-scibert} achieve only $34.57\%$ Macro-F1 score which shows that there is still plenty of room for improvement.

\section{Related Work}
 %\vspace{-2mm}
To date, several datasets exist for topic classification of scientific papers. \citet{kowsari2017HDLTex} created a hierarchically classified dataset of scientific %documents with 
papers from the Web of Science (WoS).\footnote{\Fontvim{\href{https://clarivate.com/webofsciencegroup/solutions/web-of-science/}{\texttt{https://clarivate.com/webofsciencegroup/solutions/
web-of-science/}}}} However, their hierarchy %tree in their dataset 
is only two levels deep and the size of their %largest 
dataset is $46,985$, which is much smaller than its counterpart from news source data. In addition, there are only $141$ topics in the entire hierarchy. The Cora\footnote{\Fontvim{\href{https://people.cs.umass.edu/~mccallum/data/cora-classify.tar.gz}{\texttt{https://people.cs.umass.edu/~mccallum/data/
cora-classify.tar.gz}}}} dataset introduced by \citet{cora} is also hierarchically classified with multiple labels per paper and contains about $50,000$ papers. The hierarchy varies in depth from one to three and has $79$ topics in total. However, the widely used version of Cora\footnote{\Fontvim{\href{https://linqs-data.soe.ucsc.edu/public/lbc/cora.tgz}{\texttt{https://linqs-data.soe.ucsc.edu/public/lbc/cora.tgz}}}} contains only $2,708$ papers \cite{lu2003link} and is not hierarchical. Similarly, the labeled dataset for topic classification of scientific papers from CiteSeer\footnote{\Fontvim{\href{https://linqs-data.soe.ucsc.edu/public/citeseer-mrdm05/}{\texttt{https://linqs-data.soe.ucsc.edu/public/
citeseer-mrdm05/}}}} \cite{citeseer98} is also very small in size containing only $3,312$ papers with no hierarchy over the labels. \citet{yang-etal-2018-sgm} created a dataset of $55,840$ arXiv papers where each paper is assigned multiple labels using a two-level deep topic hierarchy containing a total of $54$ topics. Similar to us, \citet{santos2009multi} proposed a multi-label hierarchical document classification dataset using the ACM category hierarchy. However, our dataset is much larger in size than this dataset (which has $\approx15,000$ documents in their experiment setting). Furthermore, the dataset by \citet{santos2009multi} is not available online and cannot be reconstructed as the ACM paper IDs are not provided.

Recently, \citet{specter2020cohan} released a dataset of $25,000$ papers collected from the Microsoft Academic Graph\footnote{\url{https://academic.microsoft.com/home}} (MAG) as part of their proposed evaluation benchmark for document level research on scientific domains. Although the papers in MAG are arranged in a five level deep hierarchy \cite{Sinha15mag}, only the level one categories ($19$ topics in total) are made available with the dataset. In contrast to the above datasets, {\bf SciHTC} has $1,233$ topics arranged in a six level deep hierarchy. The total number of papers in our dataset is $186,160$ which is significantly larger than all other datasets mentioned above. The topic hierarchy of each paper is provided by their respective authors. Since each paper in our dataset is assigned to all the topics on the path from the root to a certain topic in the hierarchy tree, our dataset can be referred to as a multi-label dataset for topic classification.

 %%\vspace{-1mm}
For multi-label classification, there are two major approaches: a) training one model to predict all the topics to which each paper belongs \cite{PengDGCNN,baker2017initializing, liu2017deep}; and b) training one-vs-all binary classifiers for each of the topics \cite{banerjee-etal-2019-hierarchical, read2009classifier}. The first approach learns to classify papers to all the relevant topics simultaneously, and hence, it is better suited to leverage the inter-label dependencies among the labels. % that occur together frequently. 
However, despite that it is simpler and more time efficient, it struggles with data imbalance \cite{banerjee-etal-2019-hierarchical}. On the other hand, the second approach gives enough flexibility to deal with the different levels of class imbalance but it is more complex and not as time efficient as the first one. In general, the second approach takes additional steps to encode the inter-label dependencies among the co-occurring labels. For example, in hierarchical classification, the parameters of the model for a child topic can be initialized with the parameters of the trained model for its parent topic \cite{kurata-etal-2016-improved, baker-korhonen-2017-initializing, banerjee-etal-2019-hierarchical}. In this work, we take both approaches and compare their performance.

Besides these approaches, another approach for hierarchical and/or multi-label classification in recent years is based on sequence-to-sequence models  \cite{yang-etal-2018-sgm, yang-etal-2019-deep}, which we explored in this work. However, 
these models failed to show satisfactory performance on our dataset. We also explored the hierarchical classification proposed by \citet{kowsari2017HDLTex} where a local  classifier is trained at every node of the hierarchy, but this model also failed to gain satisfactory performance.

\citet{onan2016ensemble} proposed the use of keywords together with traditional ensemble methods to classify scientific papers. However, since ground truth keywords were not available for the papers in their dataset, the authors explored a frequency based keyword selection, which gave the best performance. Therefore, their application of keyword extraction methods for the classification task can be seen as a feature selection method. 

Our {\bf SciHTC} dataset, in addition to being very large, multi-labeled, and hierarchical, contains the author-provided keywords for each paper. In this work, we present a thorough investigation of the usefulness of keywords for topic classification and propose a multi-task learning framework \cite{Caruana1993MultitaskLA,liu-etal-2019-multi-task} that uses keyword labeling as an auxiliary task to learn better representations for the main topic classification task. 

%\vspace{-1mm}
\section{The SciHTC Dataset}
\label{sec:dataset}
%\vspace{-2mm}
We constructed the \textbf{SciHTC} dataset from papers published by the ACM digital library, which we requested from ACM. 

Precisely, the dataset provided %to us 
by ACM has more than $300,000$ papers. However, some of these papers did not have their %title, abstract and/or 
author-specified keywords, whereas others did not have any category information. Thus, we pruned all these papers from the dataset. Finally, there were $186,160$ papers which had all the necessary attributes and the category information. The final dataset was randomly divided into train, development and test sets in an $80:10:10$ ratio. The number of examples in each set can be seen in Table \ref{data_sub}. 

\begin{table}[h!]
\small
\centering
\begin{tabular}{p{7em}  r}
\hline
\textbf{Dataset} & \textbf{Size}\\
\hline
Train  & 148,928 \\
Development  & 18,616 \\
Test & 18,616 \\
\hline
\end{tabular}
%\vspace{-2mm}
\caption{Dataset Splits.}
%\vspace{-2mm}
\label{data_sub}
\end{table}

\begin{table}[t]
\small
\centering
\begin{tabular}{p{0.35\textwidth}r}
\hline
\textbf{Category Hierarchy} & \textbf{Score}\\
\hline
\textbf{CCS $\to$ Software and its engineering $\to$ Software creation and management $\to$ Software verification and validation $\to$ Software defect analysis $\to$ Software testing and debugging} & \textbf{500}\\
CCS $\to$ Software and its engineering $\to$ Software notations and tools $\to$ General programming languages $\to$ Language features & 100\\
\hline
\textbf{Author-specified Keywords} & \\
 \hline
Dependent enumeration, data generation, invariant, pairing function, algebra, exhaustive testing, random testing, lazy evaluation, program inversion, DSL, SciFe & \\ 
\hline
\end{tabular}
\caption{Category hierarchies with different relevance scores and keywords for a paper --- both specified by the authors.}
\label{table:Category_hierarchy_example}
\end{table}

The category information of the papers in our dataset were defined based on the category hierarchy tree created by ACM. This hierarchy tree is named CCS or Computing Classification System. The root of the hierarchy tree is denoted as `CCS' and there are $13$ nodes at level 1 which represent topics such as ``Hardware,'' ``Networks,'' ``Security and Privacy,'' etc. Note that CCS itself does not represent any topic (or category). It is simply the root of the ACM hierarchy tree. There are 6 levels in the hierarchy tree apart from the root `CCS'. That is, the maximum depth among the leaf nodes in the tree is 6. However, note that the depths of different sub-branches are not uniform and there are leaf nodes in the tree with a depth less than $6$. 
\definecolor{myBlue}{RGB}{15, 44, 128}
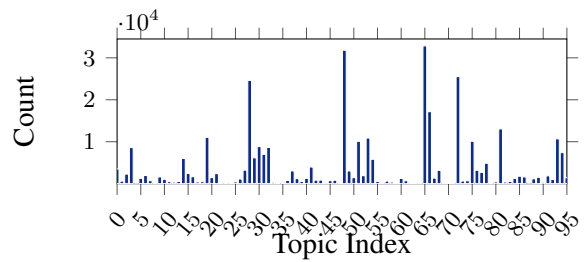
\begin{figure}[t]
\begin{flushleft}
% This file was created by tikzplotlib v0.9.6.
\begin{tikzpicture}
\begin{axis}[
width  = 7.5cm,
height = 3.5cm,
xmin=0, xmax=95,
tick align=outside,
tick pos=both,
xlabel={Topic Index},
xtick style={color=white!33.3333333333333!black},
x tick label style={rotate=50},
ylabel={Count},
ymin=0, ymax=34496.7,
ytick style={color=white!33.3333333333333!black},
ytick={10000,20000,30000, 40000},
xtick={0,5, 10,15, 20,25,30,35,40,45,50,55, 60,65,70,75, 80,85,90,95},
ticklabel style = {font=\small},
]
\draw[draw=white,fill=myBlue] (axis cs:-0.4,0) rectangle (axis cs:0.4,3518);
\draw[draw=white,fill=myBlue] (axis cs:0.6,0) rectangle (axis cs:1.4,604);
\draw[draw=white,fill=myBlue] (axis cs:1.6,0) rectangle (axis cs:2.4,2285);
\draw[draw=white,fill=myBlue] (axis cs:2.6,0) rectangle (axis cs:3.4,8608);
\draw[draw=white,fill=myBlue] (axis cs:3.6,0) rectangle (axis cs:4.4,0);
\draw[draw=white,fill=myBlue] (axis cs:4.6,0) rectangle (axis cs:5.4,1270);
\draw[draw=white,fill=myBlue] (axis cs:5.6,0) rectangle (axis cs:6.4,1947);
\draw[draw=white,fill=myBlue] (axis cs:6.6,0) rectangle (axis cs:7.4,716);
\draw[draw=white,fill=myBlue] (axis cs:7.6,0) rectangle (axis cs:8.4,12);
\draw[draw=white,fill=myBlue] (axis cs:8.6,0) rectangle (axis cs:9.4,1607);
\draw[draw=white,fill=myBlue] (axis cs:9.6,0) rectangle (axis cs:10.4,1031);
\draw[draw=white,fill=myBlue] (axis cs:10.6,0) rectangle (axis cs:11.4,477);
\draw[draw=white,fill=myBlue] (axis cs:11.6,0) rectangle (axis cs:12.4,355);
\draw[draw=white,fill=myBlue] (axis cs:12.6,0) rectangle (axis cs:13.4,541);
\draw[draw=white,fill=myBlue] (axis cs:13.6,0) rectangle (axis cs:14.4,6017);
\draw[draw=white,fill=myBlue] (axis cs:14.6,0) rectangle (axis cs:15.4,2443);
\draw[draw=white,fill=myBlue] (axis cs:15.6,0) rectangle (axis cs:16.4,1658);
\draw[draw=white,fill=myBlue] (axis cs:16.6,0) rectangle (axis cs:17.4,400);
\draw[draw=white,fill=myBlue] (axis cs:17.6,0) rectangle (axis cs:18.4,467);
\draw[draw=white,fill=myBlue] (axis cs:18.6,0) rectangle (axis cs:19.4,11036);
\draw[draw=white,fill=myBlue] (axis cs:19.6,0) rectangle (axis cs:20.4,1453);
\draw[draw=white,fill=myBlue] (axis cs:20.6,0) rectangle (axis cs:21.4,2394);
\draw[draw=white,fill=myBlue] (axis cs:21.6,0) rectangle (axis cs:22.4,171);
\draw[draw=white,fill=myBlue] (axis cs:22.6,0) rectangle (axis cs:23.4,23);
\draw[draw=white,fill=myBlue] (axis cs:23.6,0) rectangle (axis cs:24.4,34);
\draw[draw=white,fill=myBlue] (axis cs:24.6,0) rectangle (axis cs:25.4,417);
\draw[draw=white,fill=myBlue] (axis cs:25.6,0) rectangle (axis cs:26.4,1110);
\draw[draw=white,fill=myBlue] (axis cs:26.6,0) rectangle (axis cs:27.4,3227);
\draw[draw=white,fill=myBlue] (axis cs:27.6,0) rectangle (axis cs:28.4,24623);
\draw[draw=white,fill=myBlue] (axis cs:28.6,0) rectangle (axis cs:29.4,6164);
\draw[draw=white,fill=myBlue] (axis cs:29.6,0) rectangle (axis cs:30.4,8842);
\draw[draw=white,fill=myBlue] (axis cs:30.6,0) rectangle (axis cs:31.4,6985);
\draw[draw=white,fill=myBlue] (axis cs:31.6,0) rectangle (axis cs:32.4,8630);
\draw[draw=white,fill=myBlue] (axis cs:32.6,0) rectangle (axis cs:33.4,323);
\draw[draw=white,fill=myBlue] (axis cs:33.6,0) rectangle (axis cs:34.4,239);
\draw[draw=white,fill=myBlue] (axis cs:34.6,0) rectangle (axis cs:35.4,226);
\draw[draw=white,fill=myBlue] (axis cs:35.6,0) rectangle (axis cs:36.4,788);
\draw[draw=white,fill=myBlue] (axis cs:36.6,0) rectangle (axis cs:37.4,3039);
\draw[draw=white,fill=myBlue] (axis cs:37.6,0) rectangle (axis cs:38.4,1202);
\draw[draw=white,fill=myBlue] (axis cs:38.6,0) rectangle (axis cs:39.4,519);
\draw[draw=white,fill=myBlue] (axis cs:39.6,0) rectangle (axis cs:40.4,1261);
\draw[draw=white,fill=myBlue] (axis cs:40.6,0) rectangle (axis cs:41.4,3972);
\draw[draw=white,fill=myBlue] (axis cs:41.6,0) rectangle (axis cs:42.4,841);
\draw[draw=white,fill=myBlue] (axis cs:42.6,0) rectangle (axis cs:43.4,847);
\draw[draw=white,fill=myBlue] (axis cs:43.6,0) rectangle (axis cs:44.4,272);
\draw[draw=white,fill=myBlue] (axis cs:44.6,0) rectangle (axis cs:45.4,717);
\draw[draw=white,fill=myBlue] (axis cs:45.6,0) rectangle (axis cs:46.4,814);
\draw[draw=white,fill=myBlue] (axis cs:46.6,0) rectangle (axis cs:47.4,109);
\draw[draw=white,fill=myBlue] (axis cs:47.6,0) rectangle (axis cs:48.4,31810);
\draw[draw=white,fill=myBlue] (axis cs:48.6,0) rectangle (axis cs:49.4,3037);
\draw[draw=white,fill=myBlue] (axis cs:49.6,0) rectangle (axis cs:50.4,1440);
\draw[draw=white,fill=myBlue] (axis cs:50.6,0) rectangle (axis cs:51.4,10106);
\draw[draw=white,fill=myBlue] (axis cs:51.6,0) rectangle (axis cs:52.4,1906);
\draw[draw=white,fill=myBlue] (axis cs:52.6,0) rectangle (axis cs:53.4,10881);
\draw[draw=white,fill=myBlue] (axis cs:53.6,0) rectangle (axis cs:54.4,5816);
\draw[draw=white,fill=myBlue] (axis cs:54.6,0) rectangle (axis cs:55.4,514);
\draw[draw=white,fill=myBlue] (axis cs:55.6,0) rectangle (axis cs:56.4,15);
\draw[draw=white,fill=myBlue] (axis cs:56.6,0) rectangle (axis cs:57.4,616);
\draw[draw=white,fill=myBlue] (axis cs:57.6,0) rectangle (axis cs:58.4,373);
\draw[draw=white,fill=myBlue] (axis cs:58.6,0) rectangle (axis cs:59.4,44);
\draw[draw=white,fill=myBlue] (axis cs:59.6,0) rectangle (axis cs:60.4,1256);
\draw[draw=white,fill=myBlue] (axis cs:60.6,0) rectangle (axis cs:61.4,703);
\draw[draw=white,fill=myBlue] (axis cs:61.6,0) rectangle (axis cs:62.4,53);
\draw[draw=white,fill=myBlue] (axis cs:62.6,0) rectangle (axis cs:63.4,93);
\draw[draw=white,fill=myBlue] (axis cs:63.6,0) rectangle (axis cs:64.4,117);
\draw[draw=white,fill=myBlue] (axis cs:64.6,0) rectangle (axis cs:65.4,32854);
\draw[draw=white,fill=myBlue] (axis cs:65.6,0) rectangle (axis cs:66.4,17166);
\draw[draw=white,fill=myBlue] (axis cs:66.6,0) rectangle (axis cs:67.4,1326);
\draw[draw=white,fill=myBlue] (axis cs:67.6,0) rectangle (axis cs:68.4,3179);
\draw[draw=white,fill=myBlue] (axis cs:68.6,0) rectangle (axis cs:69.4,50);
\draw[draw=white,fill=myBlue] (axis cs:69.6,0) rectangle (axis cs:70.4,156);
\draw[draw=white,fill=myBlue] (axis cs:70.6,0) rectangle (axis cs:71.4,79);
\draw[draw=white,fill=myBlue] (axis cs:71.6,0) rectangle (axis cs:72.4,25529);
\draw[draw=white,fill=myBlue] (axis cs:72.6,0) rectangle (axis cs:73.4,623);
\draw[draw=white,fill=myBlue] (axis cs:73.6,0) rectangle (axis cs:74.4,698);
\draw[draw=white,fill=myBlue] (axis cs:74.6,0) rectangle (axis cs:75.4,10105);
\draw[draw=white,fill=myBlue] (axis cs:75.6,0) rectangle (axis cs:76.4,3193);
\draw[draw=white,fill=myBlue] (axis cs:76.6,0) rectangle (axis cs:77.4,2670);
\draw[draw=white,fill=myBlue] (axis cs:77.6,0) rectangle (axis cs:78.4,4875);
\draw[draw=white,fill=myBlue] (axis cs:78.6,0) rectangle (axis cs:79.4,89);
\draw[draw=white,fill=myBlue] (axis cs:79.6,0) rectangle (axis cs:80.4,421);
\draw[draw=white,fill=myBlue] (axis cs:80.6,0) rectangle (axis cs:81.4,13058);
\draw[draw=white,fill=myBlue] (axis cs:81.6,0) rectangle (axis cs:82.4,382);
\draw[draw=white,fill=myBlue] (axis cs:82.6,0) rectangle (axis cs:83.4,535);
\draw[draw=white,fill=myBlue] (axis cs:83.6,0) rectangle (axis cs:84.4,1223);
\draw[draw=white,fill=myBlue] (axis cs:84.6,0) rectangle (axis cs:85.4,1779);
\draw[draw=white,fill=myBlue] (axis cs:85.6,0) rectangle (axis cs:86.4,1612);
\draw[draw=white,fill=myBlue] (axis cs:86.6,0) rectangle (axis cs:87.4,0);
\draw[draw=white,fill=myBlue] (axis cs:87.6,0) rectangle (axis cs:88.4,1162);
\draw[draw=white,fill=myBlue] (axis cs:88.6,0) rectangle (axis cs:89.4,1519);
\draw[draw=white,fill=myBlue] (axis cs:89.6,0) rectangle (axis cs:90.4,330);
\draw[draw=white,fill=myBlue] (axis cs:90.6,0) rectangle (axis cs:91.4,1866);
\draw[draw=white,fill=myBlue] (axis cs:91.6,0) rectangle (axis cs:92.4,987);
\draw[draw=white,fill=myBlue] (axis cs:92.6,0) rectangle (axis cs:93.4,10689);
\draw[draw=white,fill=myBlue] (axis cs:93.6,0) rectangle (axis cs:94.4,7409);
\draw[draw=white,fill=myBlue] (axis cs:94.6,0) rectangle (axis cs:95.4,1428);
\draw[draw=white,fill=myBlue] (axis cs:95.6,0) rectangle (axis cs:96.4,387);
\end{axis}
\end{tikzpicture}
\caption{Number of papers in each topic up to level two of the ACM CCS hierarchy tree.}
\vspace{-5mm}
\label{paper_count_dist}
\end{flushleft}
\end{figure}

Each paper in our dataset is assigned to one or more sub-branches of the hierarchy tree by their respective authors with different depth levels and relevance scores among \{$100, 300, 500$\} with $500$ indicating the most relevant. The authors also provide a set of keywords relevant to their paper. Table \ref{table:Category_hierarchy_example} shows the assigned sub-branches and keywords of an example paper,\footnote{\url{http://dl.acm.org/citation.cfm?id=2814323}} both of them being provided by the authors. Among the author-specified sub-branches, we only consider the sub-branch with the highest relevance score for each paper. Thus, the categories in the first sub-branch (bolded line) in Table \ref{table:Category_hierarchy_example} are selected as the labels for the paper. However, considering all relevant sub-branches can present a more interesting and challenging task which can be explored in future work.  

There are $1,233$ different topics in total in our final  dataset. However, we find that the distribution of the number of papers over the topics is very imbalanced and a few topics (especially in the deeper levels of the hierarchy) had extremely low support (i.e., rare topics). Thus, for our experiments, we only consider the topics up to level 2 of the CCS hierarchy tree which had at least $100$ examples in the training set. Figure \ref{paper_count_dist} shows the number of papers in each of the $95$ topics up to level 2 of the hierarchy tree in our dataset. We also report the explicit topic distribution (i.e., topic name vs. support) in Appendix \ref{label_dist}. Note that since there are $12$ topics (among the $95$ topics up to level 2 of the hierarchy) with less than $100$ examples in the training set, we remove them and experiment with the remaining $83$ topics. Although we do not use the topics with low support in our experiments, we believe that they can be potentially useful for hierarchical topic classification of rare topics. Therefore, we make available not only the two-level hierarchy dataset used in our experiments but also all relevant topics for each paper from the six-level hierarchy tree. 

\tikzstyle{flat_model} = [rectangle,draw,fill=green!20,text width=5.5em, text centered, minimum height = 4em]

\tikzstyle{input_text} = [rectangle,draw,fill=blue!20,text width=5.5em, text centered, minimum height = 2em]

\section{Methodology}
 %\vspace{-2mm}
\label{sec:methodology}
This section describes the hierarchical and flat multi-label baselines used in our experiments (\S\ref{baseline_modeling}); after that, it introduces our simple incorporation of keywords into the models (\S\ref{sec:kws}); lastly, it presents our multi-task learning framework for topic classification (\S\ref{sec:mtl}). 

\paragraph{Problem Definition}

Let $p$ be a paper, $t$ be a topic from the set of all topics $T$, and $n$ be the number of all topics in $T$; and let ${\bf x}^p$ denote the input text and ${\bf y}^p$ denote the label vector of size $n$ corresponding to $p$. For our baseline models, ${\bf x}^p$ is a concatenation of the title and abstract of $p$. The goal is to predict the label vector ${\bf y}^p$ given ${\bf x}^p$ such that, if $p$ belongs to topic $t$, $y^{p}_{t} = 1$, and $y^{p}_{t} = 0$ otherwise, i.e.,  identify all topics relevant to $p$.

\subsection{Baseline Modeling}
%\vspace{-1mm}
\label{baseline_modeling}
We establish both flat and hierarchical classification approaches as our baselines, as discussed below.
%\vspace{-1mm}
\subsubsection{Flat Multi-Label Classification}
\label{sec:flat_multi}
We refer to the classifiers that predict all relevant topics of a paper with a single model as flat multi-label classifiers. %Precisely,
Although these models leverage the inter-label dependencies by learning to predict all relevant labels simultaneously, they do not consider the %labels being from a hierarchy. 
label hierarchy structure. In the models, all layers are shared until the last layer during training. Instead of softmax, the output layer consists of $n$ nodes, each with sigmoid activation. % where $n$ is the total number of possible labels. 
Each sigmoid output represents the probability of a topic $t$ being relevant for a paper $p$, with $t=1,\cdots,n$. The architecture %of the flat models 
is illustrated in Appendix \ref{appendix:archs}.

We use the following neural models to obtain representations of the input text: neural model Bi-LSTM \cite{hochreiter1997long}, and pre-trained language models---BERT \cite{devlin-etal-2019-bert} and SciBERT \cite{beltagy-etal-2019-scibert}. 

\paragraph{Traditional Neural Models} We use a BiLSTM based model similar to \citet{banerjee-etal-2019-hierarchical} as our traditional neural baseline. Specifically, we take three approaches to obtain a single representation of the input text from the hidden states of the Bi-LSTM and concatenate them before they are sent to the fully connected layers. These approaches are: element-wise max pool, element-wise mean pool, and an attention weighted context vector. The attention mechanism is similar to the word level attention mechanism from \citet{yang-etal-2016-hierarchical}. After the Bi-LSTM, we use one fully connected layer with ReLU activation followed by the output layer with sigmoid activation. The obtained representations are projected with $n$ weight matrices $\mathbf{W}_t \in \mathbb{R}^{d \times 1}$. We also explore a CNN based model as another neural baseline and report its performance and architectural design in Appendix \ref{cnn_model_results}.

\paragraph{Pre-trained Language Models}
We fine-tune base BERT \cite{devlin-etal-2019-bert} and SciBERT \cite{beltagy-etal-2019-scibert} using the HuggingFace\footnote{\url{https://github.com/huggingface/transformers}} transformers library. We use the ``bert-base-uncased" and ``scibert-scivocab-uncased" variants of BERT and SciBERT, respectively. Both of these language models are pre-trained on huge amounts of text. %The only difference between them is that  whereas 
While BERT is pre-trained on the BookCorpus \cite{zhu2015aligning} and Wikipedia,\footnote{\url{https://www.wikipedia.org/}} SciBERT is pre-trained exclusively on scientific documents. After getting the hidden state embedded in the \texttt{[CLS]} token from these models, we send them through a fully connected output layer to get the classification probability.
That is, we project the \texttt{[CLS]} token with $n$ weight matrices $\mathbf{W}_t \in \mathbb{R}^{d \times 1}$. The language model and classification parameters are jointly fine-tuned.

\subsubsection{Hierarchical Multi-Label Classification}
\label{sec:HMLC}
In this approach, we train $n$ one-vs-all binary classifiers. %where $n$ is the total number of possible labels. 
As with flat multi-label classification, we use both traditional neural models based architectures and pre-trained language models, which are similar to the flat architectures described in \S\ref{sec:flat_multi} with two key differences. First, the output layer no longer contains $n$ number of nodes. Since we train binary classifiers, we change the architectures by having output layers with only one node with sigmoid activation. Second, to leverage the inter-label dependencies we initialize the model parameters of a child node in the topic hierarchy tree by its parent node's trained model parameters similar to \citet{kurata-etal-2016-improved, baker-korhonen-2017-initializing, banerjee-etal-2019-hierarchical}. An illustration of this method of leveraging the topic hierarchy to learn inter-label dependencies can be seen in Appendix \ref{appendix:archs}.

\begin{figure}[t]
\begin{centering}

\vspace{2mm}
%\tikzstyle{entity1}=[thick,text width=15cm] 
\tikzstyle{entity1}=[shape=rectangle,rounded corners,draw=mynewred3,thick,minimum width  = 4.3cm, minimum height = 0.6cm] 
%\tikzstyle{entity1}=[shape=rectangle,rounded corners,draw=mynewgray,thick,minimum width  = 4.3cm,
%                                         minimum height = 0.6cm]                                          
                                         
\tikzstyle{entity2}=[shape=rectangle,rounded corners,draw=mynewblue2,thick,minimum width  = 6.5cm,
                                         minimum height = 0.6cm] 
                                         
\tikzstyle{entity3}=[draw=myblue,thick,text width=4cm, font=\footnotesize] 
\tikzstyle{entity4}=[draw=myred,thick,text width=7cm, font=\footnotesize]
\tikzstyle{entity5}=[circle,draw=mynewblue2,fill=mynewblue,thick,inner sep=1pt,minimum size=6.5pt]
\tikzstyle{entity6}=[circle,draw=mynewred2,fill=mynewred,thick,inner sep=1pt,minimum size=6.5pt]
%\tikzstyle{entity6}=[circle,draw=mynewgray,fill=gray,thick,inner sep=1pt,minimum size=6.5pt]

\tikzstyle{entity7}=[shape=rectangle,rounded corners,draw=mynewblue2,thick,minimum width  = 1.4cm,minimum height = 0.6cm] 

\tikzstyle{entity11}=[shape=rectangle,rounded corners,draw=mynewred2,thick,minimum width  = 1.4cm,minimum height = 0.6cm]

\tikzstyle{entity8}=[shape=rectangle,rounded corners,draw=mynewblue2,thick,minimum width  = 7.5cm,minimum height = 2.7cm]

\tikzstyle{entity10}=[shape=rectangle,rounded corners,fill=transformerff, draw=mynewblue2,thick,minimum width  = 6cm,minimum height = 1cm] 

\tikzstyle{entity9}=[shape=rectangle,rounded corners,fill=transformersa,draw=black,thick,minimum width  = 7cm,minimum height = 1cm]

\tikzstyle{entity12}=[circle,draw=green!30!black,fill=input,thick,inner sep=1pt,minimum size=6.5pt]
\tikzstyle{entity13}=[shape=rectangle,rounded corners,draw=green!30!black,thick,minimum width  = 1.4cm,minimum height = 0.6cm]

\tikzstyle{entity14}=[circle,draw=magenta!80!black,fill=saoutput,thick,inner sep=1pt,minimum size=6.5pt]
\tikzstyle{entity15}=[shape=rectangle,rounded corners,draw=magenta!80!black,thick,minimum width  = 1.4cm,minimum height = 0.6cm]

\tikzstyle{entity16}=[circle,draw=blue!80!black,fill=ffoutput,thick,inner sep=1pt,minimum size=6.5pt]
\tikzstyle{entity17}=[shape=rectangle,rounded corners,draw=blue!80!black,thick,minimum width  = 1.4cm,minimum height = 0.6cm]

\tikzstyle{entity18}=[shape=rectangle,rounded corners,fill=transformerff, draw=mynewblue2,thick,minimum width  = 1cm,minimum height = 1cm]

\begin{tikzpicture} [node distance=6mm][auto,thick][semithick,>=stealth']
 \pgfsetarrowsend{latex}
   
\node (z0) at (2.5, 0.9) {\texttt{[CLS]}};
\node (z0) at (4, 0.9) {Tok1};
\node (z0) at (5.5, 0.9) {Tok2};
\node (z0) at (6.7, 0.9) {$\cdots$};
\node (z0) at (8, 0.9) {TokN};

\draw[mynewgray,thick] (2.45,1.2) -- (2.45,1.7); 
\draw[mynewgray,thick] (3.95,1.2) -- (3.95,1.7); 
\draw[mynewgray,thick] (5.45,1.2) -- (5.45,1.7); 
\draw[mynewgray,thick] (8,1.2) -- (8,1.7);

\node[entity13] (z4) at (2.45,2) {}; 
\node[entity13] (z4) at (3.95,2) {}; 
\node[entity13] (z4) at (5.45,2) {}; 
\node[entity13] (z4) at (8,2) {}; 
\node (z0) at (6.7, 2) {$\cdots$};

\node[entity12] (z5) at (2,2) {};
\node[entity12] (z6) at (2.3,2) {};
\node[entity12] (z6) at (2.6,2) {};
\node[entity12] (z6) at (2.9,2) {};

\node[entity12] (z5) at (3.5,2) {};
\node[entity12] (z6) at (3.8,2) {};
\node[entity12] (z6) at (4.1,2) {};
\node[entity12] (z6) at (4.4,2) {};

\node[entity12] (z5) at (5,2) {};
\node[entity12] (z6) at (5.3,2) {};
\node[entity12] (z6) at (5.6,2) {};
\node[entity12] (z6) at (5.9,2) {};

\node[entity12] (z5) at (7.55,2) {};
\node[entity12] (z6) at (7.85,2) {};
\node[entity12] (z6) at (8.15,2) {};
\node[entity12] (z6) at (8.45,2) {};

%next set of arrows
\draw[mynewgray,thick] (2.45,2.4) -- (2.45,2.9); 
\draw[mynewgray,thick] (3.95,2.4) -- (3.95,2.9); 
\draw[mynewgray,thick] (5.45,2.4) -- (5.45,2.9); 
\draw[mynewgray,thick] (8,2.4) -- (8,2.9);

 \node[entity8] (z4) at (5.2,4) {}; 

\node[entity9] (z4) at (5.2,3.4) {BERT};

\tikzstyle{entity55}=[shape=rectangle,rounded corners,draw=red,thick,dashed,minimum width  = 1.8cm,minimum height = 2.5cm] 

\tikzstyle{entity65}=[shape=rectangle,rounded corners,draw=cyan,thick,dashed,minimum width  = 5.6cm,minimum height = 2.5cm]

 \node[entity55] (z4) at (2.3,6.7) {}; 
 \node[entity65] (z4) at (6.1,6.7) {};

\node[entity17] (z4) at (2.45,4.9) {}; 
\node[entity17] (z4) at (3.95,4.9) {}; 
\node[entity17] (z4) at (5.45,4.9) {}; 
\node[entity17] (z4) at (8,4.9) {}; 

\node[entity16] (z5) at (2,4.9) {};
\node[entity16] (z6) at (2.3,4.9) {};
\node[entity16] (z6) at (2.6,4.9) {};
\node[entity16] (z6) at (2.9,4.9) {};

\node[entity16] (z5) at (3.5,4.9) {};
\node[entity16] (z6) at (3.8,4.9) {};
\node[entity16] (z6) at (4.1,4.9) {};
\node[entity16] (z6) at (4.4,4.9) {};

\node[entity16] (z5) at (5,4.9) {};
\node[entity16] (z6) at (5.3,4.9) {};
\node[entity16] (z6) at (5.6,4.9) {};
\node[entity16] (z6) at (5.9,4.9) {};

\node[entity16] (z5) at (7.55,4.9) {};
\node[entity16] (z6) at (7.85,4.9) {};
\node[entity16] (z6) at (8.15,4.9) {};
\node[entity16] (z6) at (8.45,4.9) {};

\node (z0) at (6.7, 4.9) {$\cdots$};

\draw[mynewgray,thick] (2.45,5.2) -- (2.45,5.7); 
\draw[mynewgray,thick] (3.95,5.2) -- (3.95,5.7); 
\draw[mynewgray,thick] (5.45,5.2) -- (5.45,5.7); 
\draw[mynewgray,thick] (8,5.2) -- (8,5.7);

\draw[mynewgray,thick] (2.45,4) -- (2.45,4.5); 
\draw[mynewgray,thick] (3.95,4) -- (3.95,4.5); 
\draw[mynewgray,thick] (5.45,4) -- (5.45,4.5); 
\draw[mynewgray,thick] (8,4) -- (8,4.5);

\draw[mynewgray,thick] (2.45,6.7) -- (2.45,7.2); 
\draw[mynewgray,thick] (3.95,6.7) -- (3.95,7.2); 
\draw[mynewgray,thick] (5.45,6.7) -- (5.45,7.2); 
\draw[mynewgray,thick] (8,6.7) -- (8,7.2);

 \node (z30) at (2.3,7.4) {\Fontvim{Pred. topic}}; 
 \node (z30) at (3.95,7.4) {\Fontvim{0}};
 \node (z30) at (5.45,7.4) {\Fontvim{1}};
 \node (z30) at (8,7.4) {\Fontvim{0}};
 
  \node (z30) at (2.3,7.75) {\Fontvim{Topic cls. / $\mathcal{L}_1(\theta)$}}; 
  \node (z30) at (6.3,7.75) {\Fontvim{Keyword labeling / $\mathcal{L}_2(\theta)$}};
  
  \draw[mynewgray,thick] (6,8) -- (6,8.6);
  \draw[mynewgray,thick] (2.5,8) -- (2.5,8.6);

\tikzstyle{entity19}=[shape=rectangle,rounded corners,fill=red!50!white, draw=red,thick,minimum width  = 1cm,minimum height = 1cm]

\node[entity19] (z4) at (2.3,6.2) {FF};
\node[entity18] (z4) at (3.9,6.2) {FF};
\node[entity18] (z4) at (5.5,6.2) {FF};
\node[entity18] (z4) at (8,6.2) {FF};
\node (z0) at (6.7, 6.2) {$\cdots$};

\tikzstyle{entity50}=[shape=rectangle,rounded corners,fill=gray!40!white, draw=black,thick,minimum width  = 7.5cm,minimum height = 0.7cm] 
 \node[entity50] (z4) at (5.2,9) {$\mathcal{L}(\theta) =  \alpha\mathcal{L}_1(\theta) + \beta\mathcal{L}_2(\theta)$};

\end{tikzpicture}

\end{centering}
   \caption{The architecture of our proposed multi-task learning model using BERT as the encoder. The model jointly learns two tasks: topic classification and keyword labeling. The shared layers are at the bottom whereas the task-specific layers are at the top. 
   }
    \label{Arch}
     \vspace{-4mm}
\end{figure}
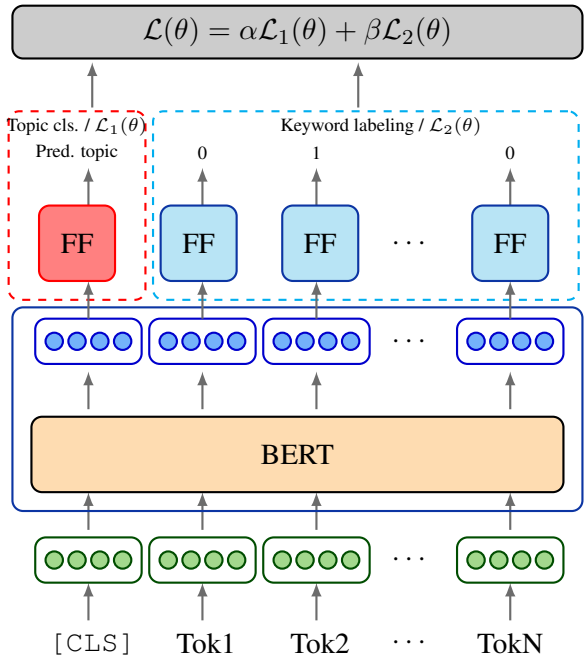

%\vspace{-2mm}
\subsection{Incorporating Keywords}
%\vspace{-1mm}
\label{sec:kws}
We aim to improve upon the baseline models described above by incorporating the keywords specified by the authors of every paper into the model. 
The keywords of a paper can provide fine-grained topical information specific to a paper and at the same time are indicative about the general (coarse-grained) topics of the paper. Thus, the keywords can be seen as a bridge between the general topics of a paper and the fine details available in it (see Table \ref{table:Category_hierarchy_example} for examples of general topics and keywords of a paper for the fine nuances of each).

We incorporate the keywords by a simple concatenation approach. The input text ${\bf x}^p$ is extended with the keywords ${\bf k}^p$ specified by the authors of $p$.
\begin{equation}
\label{equation:incorporating_keywords}
    {\bf x}^p := [{\bf x}^p,{\bf k}^p]
\end{equation}
We use the same network architectures as in \S\ref{baseline_modeling}, in both flat and hierarchical settings.

Although this approach strikes by its simplicity and, as we will see in experiments, improves over the baselines in \S\ref{baseline_modeling} that use only the title and abstract as input, it is often the case that at test time the keywords of a paper are not always available, which affects the results. Our aim is to build models that are robust enough even in the absence of keywords for papers at test time. Our proposal is to explicitly teach the model to learn to recognize the keywords in the paper that are indicative of its topics, using multi-task learning.  

\subsection{Multi-Task Learning with Keywords}
\label{sec:mtl}
%\vspace{-1mm}
We propose a neural multi-task learning framework for topic classification of scientific papers where the main task of topic classification is informed by the keyword labeling auxiliary task which aims to identify the keywords in the input text.  %We first describe our keyword identification task and then present our multi-task learning framework. 

%%\vspace{-1.5mm}
\paragraph{Keyword Labeling} Given an input sequence ${\bf x}^p=\{{\bf x}_1^p, \cdots, {\bf x}_N^p\}$ (e.g., title and abstract), the objective is to  predict a sequence of labels ${\bf z}=\{z_1,\cdots,z_N\}$, where each label $z_i$ is 1 (a keyword) or 0 (not a keyword), i.e., predict whether a word in the sequence is a keyword or not. During training, we do an exact match of the tokenized author-specified keywords in the tokenized input text (title+abstract) and set the keyword label $z_i$ as 1 for the positions where we find a match in the input text and 0 otherwise.

\paragraph{Multi-Task Learning Model} The architecture of our multi-task learning model is shown in Figure \ref{Arch}. It jointly learns two tasks: topic classification and keyword labeling. As can be seen from the figure, the model has shared layers across both tasks at the bottom and task-specific layers at the top. In the figure, we show BERT as the encoder to avoid clutter, but in experiments we use all encoders described in  \S\ref{baseline_modeling}. 
 
\setlength{\parskip}{0.6em}
\noindent {\em Shared layers.} The input of the model is the sequence ${\bf x}^p$ of $N$ words. These words are first mapped into word embedding vectors (e.g., by summing word and positional embeddings in BERT), which are then fed into the encoder block that produces a sequence of contextual embeddings (one for each input token, including the \texttt{[CLS]} token in the transformer-based models).

\setlength{\parskip}{0.6em}
\noindent {\em Task-specific layers.} There are two task-specific output layers. The topic classification output layer works the same as discussed in \S\ref{baseline_modeling}. On the other hand, the output layer for keywords labeling consists of a fully connected layer with sigmoid activation which predicts whether a word is a keyword or not by using each token contextual embeddding. 

 \vspace{-3mm}
\paragraph{Training}
The model is optimized based on both tasks. During training, two losses are calculated for the two tasks (main and auxiliary) and they are combined together as a sum. This summed loss is used to optimize the model parameters. 

The overall loss $\mathcal{L}(\theta)$ in our model is as follows:
\begin{equation}
\mathcal{L}(\theta) = \alpha\mathcal{L}_1(\theta) + \beta \mathcal{L}_2(\theta)
\end{equation}
where $\mathcal{L}_1(\theta)$ is the loss for topic classification and $\mathcal{L}_2(\theta)$ is the loss for keyword labeling. $\alpha$ and $\beta$ are hyperparameters to scale the losses with different weights. 

\vspace{-3mm}
\section{Experiments and Results}
\vspace{-4mm}
\label{section:experiments_results}
We perform the following experiments. First, we study the difficulty of classifying topics for scientific papers from our dataset in comparison with related datasets ({\S \ref{sec:SciHTCvOthers}}). Second, we show the impact of the hierarchy of topics on models' performance and how incorporating keywords can help improve the performance further ({\S \ref{sec:hierarchyAndKeywordsImpact}}). Third, we evaluate the performance of our proposed multi-task learning approach ({\S \ref{sec:MultiTask}}). Implementation details are reported in Appendix \ref{experimental_Settings}.

\vspace{-2mm}
\subsection{SciHTC vs. Related Datasets}
 \vspace{-3mm}
\label{sec:SciHTCvOthers}

We contrast {\bf SciHTC} with three related datasets: Cora Research Paper Classification %\footnote{\href{https://people.cs.umass.edu/~mccallum/data.html}{https://people.cs.umass.edu/~mccallum/data.html}} 
\cite{cora}, WoS-46985 %\footnote{\href{https://data.mendeley.com/datasets/9rw3vkcfy4/6}{https://data.mendeley.com/datasets/9rw3vkcfy4/6}} 
\cite{kowsari2017HDLTex} and the Microsoft Academic Graph (MAG) dataset released by \citet{specter2020cohan}.

\begin{table}
\small
\centering
\begin{tabular}{lrrcc}
\hline
{Dataset} & \textbf{$n$} & Size & {F-BiLSTM}  & {HR-BiLSTM}\\
\hline
MAG & 19 & 25,000 & $70.52\%$ & $-$\\
Cora & 79 & 50,000 & $45.71\%$  & $51.79\%$\\
WoS & 141 & 46,985 & $51.99\%$ & $60.12\%$\\
SciHTC & 83 & 186,160 & $24.78\%$ & $28.54\%$\\
\hline
\end{tabular}
\caption{Number of topics $n$, dataset size, and Macro F1 of flat (F) and hierarchical (HR) Bi-LSTM.}% models on different datasets.}
\label{table:performance_comparaison_other_datasets}
 \vspace{-3mm}
\end{table}
 %%\vspace{-1mm}

The WoS dataset does not have the titles of the papers. Therefore, only the abstracts are used as the input sequence. For the Cora and MAG datasets as well as our \textbf{SciHTC} dataset, we use both title and abstract as the input sequence. We use the train, test and validation splits released by \citet{specter2020cohan} for the MAG dataset but we split the other two datasets (Cora and WoS) in a 80:10:10 ratio similar to ours because they did not have author defined splits. For this experiment, our goal was to compare the degree of difficulty of our dataset with respect to the other related datasets. We thus choose to experiment with both flat and hierarchical baseline Bi-LSTM models with 300D Glove embeddings. On the MAG dataset, we only report the Flat Bi-LSTM performance since only level 1 categories are made available by the authors (with no hierarchy information). 
We experiment with the categories up to level 2 of the hierarchy tree for the other datasets. Table \ref{table:performance_comparaison_other_datasets} shows the Macro F1 scores of these models on the four datasets along with the number of topics in each dataset and the size of each dataset. We find that:
 \vspace{-3mm}
\paragraph{SciHTC is consistently more challenging compared with related datasets.}

\begin{table*}[t!]
\centering
\small
\begin{tabular}{ll*{4}{c}}
\hline

\multicolumn{1}{c}{\textbf{Approach}} & & \textbf{Precision} & \textbf{Recall} & \textbf{Micro-F1} & \textbf{Macro-F1} \\
\hline

\hline
Flat-BiLSTM & w/o \mbox{} KW & 24.02 $\pm$ 0.37 & 28.84 $\pm$ 0.90 & 46.00 $\pm$ 0.37 & 25.36 $\pm$ 0.50 \\

Flat-BERT & w/o \mbox{} KW & 28.15 $\pm$ 0.32 & 34.65 $\pm$ 0.47 & 50.03 $\pm$ 0.14 & 30.01 $\pm$ 0.11 \\

Flat-SciBERT & w/o \mbox{} KW & 29.38 $\pm$ 0.20 & 35.92 $\pm$ 0.51 & 51.23 $\pm$ 0.30 & 31.30 $\pm$ 0.27 \\
\hline
HR-BiLSTM & w/o \mbox{} KW & 26.69 $\pm$ 0.41 & 33.38 $\pm$ 0.60 & 47.36$^{*}$ $\pm$ 0.10 & 28.73$^{*}$ $\pm$ 0.25 \\

HR-BERT & w/o \mbox{} KW & 31.06 $\pm$ 0.36 & 36.44 $\pm$ 1.94 & 51.23$^{*}$ $\pm$ 0.24 & 32.20$^{*}$ $\pm$ 0.77 \\

HR-SciBERT & w/o \mbox{} KW & 31.19 $\pm$ 0.63 & 38.27 $\pm$ 0.19 & 52.16$^{*}$ $\pm$ 0.11 & 33.13$^{*}$ $\pm$ 0.26\\
\hline

Flat-BiLSTM & with KW & 26.28 $\pm$ 0.87 & 32.02 $\pm$ 0.56 & 48.01$^{\#}$ $\pm$ 0.43 & 27.53$^{\#}$ $\pm$ 0.32 \\

Flat-BERT & with KW & 28.85 $\pm$ 0.67 & 35.41 $\pm$ 0.12 & 50.97$^{\#}$ $\pm$ 0.26 & 30.91 $\pm$ 0.38 \\

Flat-SciBERT & with KW & 30.85 $\pm$ 0.21 & 36.27 $\pm$ 0.21 & 52.01 $\pm$ 0.72 & 32.47$^{\#}$ $\pm$ 0.06 \\

\hline
HR-BiLSTM & with KW & 28.39 $\pm$ 0.07 & 35.18 $\pm$ 1.34 & 49.07$^{*\#}$ $\pm$ 0.02 & 30.59$^{*\#}$ $\pm$ 0.44 \\

HR-BERT & with KW & 31.99 $\pm$ 0.09 & 37.82 $\pm$ 0.07 & 52.26$^{*\#}$ $\pm$ 0.15 & 33.64$^{*}$ $\pm$ 0.26 \\

HR-SciBERT & with KW & 32.88 $\pm$ 0.47 & 39.37 $\pm$ 0.50 & \textbf{53.17$^{*\#}$ $\pm$ 0.18} & \textbf{34.57$^{*\#}$ $\pm$ 0.05} \\

\hline

\hline
\end{tabular}
\vspace{-1mm}
\caption{Performance comparison between models which use keywords vs. models which do not use keywords. HR - hierarchical, KW - keywords. Best results (Micro-F1 and Macro-F1) are shown in bold. Here, $^*$ and $^\#$ indicate statistically significant improvements by the HR models over their flat counterparts ($^*$) and with KW models over their w/o KW counterparts ($^\#$), respectively, according to a paired T-test with significance level $\alpha = 0.05$.}
 %\vspace{-2mm}
\label{table:with_vs_without_keywords}
\end{table*}
As we can see from Table \ref{table:performance_comparaison_other_datasets}, both models (flat and hierarchical) show a much lower performance on \textbf{SciHTC} compared with the other  datasets. %Therefore, 
It is thus evident that the degree of difficulty is much higher on our dataset, making it a more challenging benchmark for evaluation.

\begin{table*}[t!]
\centering
\small
\begin{tabular}{ll*{4}{c}}
\hline
\multicolumn{1}{c}{\textbf{Approach}} & & \textbf{Precision} & \textbf{Recall} & \textbf{Micro-F1} & \textbf{Macro-F1} \\

\hline
Flat-BiLSTM & with KW\textsuperscript{tr} & 24.10 $\pm$ 1.01 & 27.99 $\pm$ 0.79 & 45.48 $\pm$ 0.54 & 25.04 $\pm$ 0.67 \\

Flat-BERT & with KW\textsuperscript{tr} & 28.01 $\pm$ 1.17 & 34.47 $\pm$ 1.50 & 49.26 $\pm$ 0.40 & 29.44 $\pm$ 0.38 \\

Flat-SciBERT & with KW\textsuperscript{tr} & 29.73 $\pm$ 0.36 & 34.12 $\pm$ 0.31 & 50.42 $\pm$ 0.60 & 30.96 $\pm$ 0.09 \\

\hline
HR-BiLSTM & with KW\textsuperscript{tr} & 27.91 $\pm$ 0.08 & 30.20 $\pm$ 1.03 & 47.16 $\pm$ 0.04 & 28.18 $\pm$ 0.32 \\

HR-BERT & with KW\textsuperscript{tr} & 31.76 $\pm$ 0.27 & 34.65 $\pm$ 0.04 & 51.04 $\pm$ 0.01 & 32.06 $\pm$ 0.14 \\

HR-SciBERT & with KW\textsuperscript{tr} & 32.13 $\pm$ 0.37 & 35.91 $\pm$ 0.73 & 51.93 $\pm$ 0.23 & 32.67 $\pm$ 0.25\\

\hline

Flat-BiLSTM & Multi-Task & 23.90 $\pm$ 0.39 & 29.88 $\pm$ 0.57 & 46.18 $\pm$ 0.91 & 25.68 $\pm$ 0.24 \\

Flat-BERT & Multi-Task & 28.16 $\pm$ 0.94 & 34.49 $\pm$ 0.56 & 51.03$^*$ $\pm$ 0.12 & 30.06 $\pm$ 0.06 \\

Flat-SciBERT & Multi-Task & 31.21 $\pm$ 1.18 & 36.63 $\pm$ 0.79 & 52.24 $\pm$ 0.07 & 32.31$^*$ $\pm$ 0.28 \\

\hline

HR-BiLSTM & Multi-Task & 27.04 $\pm$ 0.32 & 33.66 $\pm$ 0.35 & 47.57$^*$ $\pm$ 0.23 & 29.11$^*$ $\pm$ 0.26 \\

HR-BERT & Multi-Task & 31.06 $\pm$ 0.07 & 37.20 $\pm$ 1.55 & 51.42$^*$ $\pm$ 0.01 & 32.32 $\pm$ 0.59 \\

HR-SciBERT & Multi-Task & 32.52 $\pm$ 0.89 & 38.01 $\pm$ 0.33 & \textbf{52.48 $\pm$ 0.02}  & \textbf{33.78$^*$ $\pm$ 0.18} \\

\hline
\end{tabular}
 \vspace{-1mm}
\caption{Performance comparison between models which use keywords during training by concatenating them using Eq. \ref{equation:incorporating_keywords} but not during testing vs.  models trained using multi-task learning which also do not use keywords at test time. The superscript \textsuperscript{tr} on KW\textsuperscript{tr} indicates that the keywords were concatenated only during training. Best results (Micro-F1 and Macro-F1) are shown in bold. Here, $^*$ indicates statistically significant improvements of the multi-task models over the KW\textsuperscript{tr} models according to a paired T-test with significance level $\alpha = 0.05$.} 
 \vspace{-2mm}
\label{table:results_multi_task}
\end{table*}

\vspace{-2mm}
An inspection into the categories of the related datasets revealed that these categories are more easily distinguishable from each other. For example, the categories in WoS and MAG cover broad fields of science with small overlaps between them. They range from Psychology, Medical Science, Biochemistry to Mechanical Engineering, Civil Engineering, and Computer Science. The vocabularies used in these categories/fields of science are quite different from each other and thus, the models learn to differentiate between them more easily. On the other hand, in our dataset, all papers are from the ACM digital library which are related to Computer Science and are classified to more fine-grained topics than the ones from the above datasets. Examples of topics from our dataset include Network Architectures, Network Protocols, Software Organization and Properties, Software Creation and Management, Cryptography, Systems Security, etc. Therefore, it is more difficult for the models to learn and recognize the fine differences in order to classify the topics correctly resulting in lower performance compared to the other datasets.

\vspace{-3.5mm}
\subsection{Impact of Hierarchy and Keywords}
\vspace{-3mm}
\label{sec:hierarchyAndKeywordsImpact}
Next, we explore the usefulness of the hierarchy of topics and  keywords for topic classification on {\bf SciHTC}. We experiment with all of our baseline models (flat and hierarchical) described in \S\ref{baseline_modeling} and with the incorporation of keywords described in \S\ref{sec:kws}. Precisely, each model is evaluated twice: first using only the input sequence (title+abstract) without the keywords and second by concatenating the input sequence with the keywords as in Eq. \ref{equation:incorporating_keywords}. We run each experiment three times and report their average and standard deviation in Table \ref{table:with_vs_without_keywords}. As we can see, the standard deviations of the performance scores shown by the models are very low. This illustrates that the models are stable and easily reproducible. We make the following observations:

\vspace{-4mm}
\paragraph{The hierarchy of topics improves topic classification.}
We can observe from Table \ref{table:with_vs_without_keywords} that all hierarchical models show a substantially higher performance than their flat counterparts regardless of using keywords or not. Given that the flat models learn to predict all relevant labels for each document simultaneously, it is possible for them to learn inter-label dependencies to some extent. However, due to the unavailability of the label hierarchy, the nature of the inter-label dependencies is not specified for the flat models. As a result, they can learn some spurious patterns among the labels which are harmful for overall performance. In contrast, for the hierarchical models we can specify how the inter-label dependencies should be learned (by initializing a child's model with its parent's model) which helps improve the performance as we can see in our results.

\vspace{-2mm}
\paragraph{Incorporating keywords brings further improvements.}
From Table \ref{table:with_vs_without_keywords}, we can also see that the performance of all of our baseline models increases when keywords are incorporated in the input sequence. These results illustrate that indeed, the fine-grained topical information provided by the keywords of each paper is beneficial for predicting its categorical labels (and thus capture an add-up effect for identifying the relevant coarser topics). Moreover, keywords can provide additional information which is unavailable in the title and the abstract but is relevant of the rest of the content and indicative of the topic of the paper. This additional information also helps the models to make better predictions.

\vspace{-3.5mm}
\paragraph{Transformer-based models consistently outperform Bi-LSTM models and SciBERT performs best.} 
BERT and SciBERT show strong performance across all settings (hierarchical vs. flat and with keywords vs. without) in comparison with the BiLSTM models. Interestingly, even the \textit{flat} transformer based models outperform all BiLSTM based models (including hierarchical). We believe that this is because BERT and SciBERT are pre-trained on a large amount of text. Therefore, they are able to learn better representations of the words in the input text. Comparing the two transformer based models (BERT and SciBERT), SciBERT shows the better performance. We hypothesize that this is because SciBERT's vocabulary is more relevant to the scientific domain and it is pre-trained exclusively on scientific documents. Hence, it has a better knowledge about the language used in scientific documents.

\vspace{-3mm}
\subsection{Multi-task Learning Performance}
\vspace{-2mm}
\label{sec:MultiTask}
The results in Table \ref{table:with_vs_without_keywords} show that the keywords are useful for topic classification but it is assumed that these keywords are available for papers not only during training but also at test time.
%To answer this question, 
However, often at test time the keywords of a paper are not available. We turn now to the evaluation of models when keywords are not available at test time. We compare our multi-task approach  ({\S\ref{sec:mtl}}) with the models trained with concatenating the keywords in the input sequence (during training) but tested only on the input sequence without keywords. The motivation behind this comparison is to understand the difference in performance of the models which leverage keywords during training in a manner different from our multi-task models but not at test time (same as the models trained with our multi-task approach). These results are shown in Table \ref{table:results_multi_task}. We found that:

\vspace{-2.5mm}
\paragraph{Multi-task learning effectively makes use of keywords for topic classification.}
A first observation is that {\em not making use} of gold (author-specified) keywords at test time (but only during training KW$^{tr}$ through concatenation using Eq. \ref{equation:incorporating_keywords}) decreases performance (see Table \ref{table:with_vs_without_keywords} bottom half and Table \ref{table:results_multi_task} top half). Remarkably, the multi-tasking models (which also do not use gold keywords at test time) are better at classifying the topics than the models that use keywords only during training through concatenation. %them with the title and abstract. 
In addition, comparing the models that do not use keywords at all and the multi-task models (top half of Table \ref{table:with_vs_without_keywords} and bottom half of Table \ref{table:results_multi_task}), we can see that the multi-task models perform better. Furthermore, the performance of the multi-tasking models is only slightly worse compared with that of the models that use gold (author-specified) keywords both during train and test (see bottom halves of Tables \ref{table:with_vs_without_keywords} and \ref{table:results_multi_task}). These results indicate that the models trained with our multi-task learning approach learn better representations of the input text which help improve the classification performance, thereby harnessing the usefulness of author-specified keywords even in their absence at test time.

%\newpage
\vspace{-3mm}
\section{Analysis and Discussion}
\vspace{-3mm}
From our experiments, it is evident that all of our hierarchical baselines can outperform their flat counterparts. But it is not clear whether the performance gain comes from using the hierarchy to better learn the parent-child dependency or it is because we allow the models to focus on each class individually
%are emphasizing more on each of the classes 
by training one-vs-all binary classifiers in our hierarchical setting as opposed to one flat model for all the classes. In addition, our experiments also show that keywords can be used in multiple ways to improve topic classification performance. However, it is unclear whether or not keywords {\em by themselves} can achieve the optimal performance. Thus, we analyze our models in these aspects with the following experiments. 

\begin{table}
\centering
\small
\begin{tabular}{lcc}
\hline
\textbf{Approach} & \textbf{Micro F1}  & \textbf{Macro F1}\\
\hline
HR-SciBERT with KW & $53.04\%$ & $34.61\%$\\
$n$-Binary-SciBERT with KW & $53.01\%$ & $32.44\%$\\
\hline
\end{tabular}
%\vspace{-1mm}
\caption{Comparison of performance between hierarchical SciBERT and $n$-binary SciBERT models which do not learn the parent-child relationships.}
\vspace{-4mm}
\label{table:performance_comparaison_HRvsbinarys}
\end{table}

\vspace{-4mm}
\paragraph{Hierarchical vs. $n$-Binary} We conduct an experiment with SciBERT where we train a binary classifier for each class similar to the hierarchical SciBERT model but do not initialize it with its parent's model parameters, i.e., we do not make use of the topic hierarchy. We compare the performance of this \textit{n-binary-SciBERT} model with \textit{HR-SciBERT} model in Table \ref{table:performance_comparaison_HRvsbinarys}.

\vspace{-1.5mm}
We can see that the non-hierarchical approach with $n$ binary models has more than $2$ percentage points lower Macro F1. The performance of deep learning models depends partly on how their parameters are initialized \cite{bengio2017deep}. For the \textit{n-binary} approach, since we initialize the model parameters for each class with a SciBERT model pre-trained on unsupervised data, it is forced to learn to distinguish between the examples belonging to this class and the examples from all other classes from scratch. In contrast, when the model parameters for a node in the topic hierarchy are initialized with its parent node's trained model (for HR models), we start with a model which already knows a superset of the distinct characteristics of the documents belonging to this node (i.e., the characteristics of the papers which belong to its parent node). In other words, the model does not need to be trained to classify from scratch. Therefore, the hierarchical classification setup acts as a better parameter initialization strategy which leads to a better performance.

\vspace{-3.5mm}
\paragraph{With Keywords vs. Only Keywords}We experiment with flat BiLSTM, BERT and SciBERT models with only keywords as the input. A comparison of these \textit{only keywords models} with the models which use title, abstract and keywords can be seen in Table \ref{table:performance_comparaison_only_keywords}. We can see a decline of $\approx12\%$, $\approx8\%$ and $\approx5\%$ in Macro F1 for BiLSTM, BERT and SciBERT, respectively, when only keywords are used as the input. Therefore, we can conclude that keywords are useful in topic classification but that usefulness is evident when other sources of input are also available.
\begin{table}
\centering
\small
\begin{tabular}{lcc}
\hline
\textbf{Approach} & \textbf{Micro F1}  & \textbf{Macro F1}\\
\hline
Flat-BiLSTM with KW & $47.78\%$ & $27.40\%$\\
Flat-BiLSTM only KW & $33.63\%$ & $15.87\%$\\
\hline
Flat-BERT with KW & $51.84\%$ & $30.64\%$\\
Flat-BERT only KW & $48.14\%$ & $22.95\%$\\
\hline
Flat-SciBERT with KW & $52.22\%$ & $32.41\%$\\
Flat-SciBERT only KW & $49.38\%$ & $27.18\%$\\

\hline
\end{tabular}
%\vspace{-3mm}
\caption{Comparison of performance of flat BiLSTM, BERT and SciBERT trained and tested with only keywords (no title and abstract) vs. trained and tested with title+abstract+keywords.}
\label{table:performance_comparaison_only_keywords}
 \vspace{-4mm}
\end{table}

 \vspace{-3mm}
\section{Conclusion}
\vspace{-3.5mm}
In this paper, we introduce {\bf SciHTC}, a new dataset for hierarchical multi-label classification of scientific papers and establish several strong baselines. Our experiments show that {\bf SciHTC} presents a challenging benchmark and that keywords can play a vital role in improving the classification performance. Moreover, we propose a multi-task learning framework for topic classification and keyword labeling which improves the performance over the models that do not have keywords available at test time. We believe that {\bf SciHTC} is large enough for fostering research on designing efficient models and will be a valuable resource for hierarchical multi-label classification. In our future work, we will explore novel approaches to further exploit the topic hierarchy and adopt few-shot and zero-shot learning methods to handle the extremely rare categories. We will also work on creating datasets from other domains of science with similar characteristics as SciHTC to allow further explorations.

\vspace{-3mm}
\section{Limitations}
\vspace{-4mm}
One limitation of our proposed dataset could potentially be that all of our papers are from the computer science domain and therefore, it does not provide coverage to papers from other scientific areas. However, we see this as a strength of our dataset rather than a weakness. There are other datasets already available which cover a diverse range of scientific areas (e.g., WoS). In contrast, we address the lack of a resource which can be used to study hierarchical classification among fine-grained topics, with potential confusable classes. SciHTC can be used as a benchmark for judging the models’ ability in distinguishing very subtle differences among documents belonging to closely related but different topics which will lead to development of more sophisticated models.

\vspace{-3mm}
\section*{Acknowledgements} 
\vspace{-3mm}
This research is supported in part by NSF CAREER
award \#1802358, NSF CRI award \#1823292, NSF IIS award \#2107518, and UIC Discovery Partners Institute (DPI) award.
Any opinions, findings, and conclusions expressed
here are those of the authors and do not necessarily
reflect the views of NSF and DPI. We thank AWS for computational resources. We also thank our anonymous
reviewers for their constructive feedback.

\bibliography{emnlp2020}
\bibliographystyle{acl_natbib}
\clearpage
\begin{table*}[t!]
\centering
\small
\begin{tabular}{ll*{4}{c}}
\hline

\multicolumn{1}{c}{\textbf{Approach}} & & \textbf{Precision} & \textbf{Recall} & \textbf{Micro-F1} & \textbf{Macro-F1} \\
\hline

\hline

Flat-XML-CNN & w/o \mbox{} KW & 23.75 $\pm$ 1.08 & 28.77 $\pm$ 0.50 & 45.18 $\pm$ 0.07 & 24.73 $\pm$ 0.21 \\

Flat-XML-CNN & with KW & 25.73 $\pm$ 0.93 & 31.66 $\pm$ 1.32 & 48.22 $\pm$ 0.29 & 27.31 $\pm$ 0.56 \\

Flat-XML-CNN & with KW\textsuperscript{tr} & 23.97 $\pm$ 0.34 & 27.68 $\pm$ 1.47 & 45.54 $\pm$ 0.44 & 24.84 $\pm$ 0.58 \\

Flat-XML-CNN & Multi-Task & 22.13 $\pm$ 1.04 & 26.97 $\pm$ 0.76 & 44.91 $\pm$ 0.40 & 23.13 $\pm$ 0.17 \\

\hline

HR-XML-CNN & w/o \mbox{} KW & 24.11 $\pm$ 0.53 & 30.78 $\pm$ 1.55 & 45.12 $\pm$ 0.04 & 26.25 $\pm$ 0.71 \\

HR-XML-CNN & with KW & 24.81 $\pm$ 0.37 & 33.01 $\pm$ 0.22 & 46.15 $\pm$ 0.17 & 27.58 $\pm$ 0.14 \\

HR-XML-CNN & with KW\textsuperscript{tr} & 24.75 $\pm$ 0.31 & 27.57 $\pm$ 0.59 & 44.01 $\pm$ 0.54 & 25.24 $\pm$ 0.02 \\

HR-XML-CNN & Multi-Task & 23.70 $\pm$ 0.67 & 31.97 $\pm$ 0.18 & 44.79 $\pm$ 0.37 & 26.40 $\pm$ 0.42 \\

\hline

\hline
\end{tabular}
\caption{Comparison of performance among different CNN based models. Here, HR - hierarchical, KW - keywords. The superscript \textsuperscript{tr} on KW\textsuperscript{tr} indicates that the keywords were used only during training.}
\vspace{-6mm}
\label{table:cnn_results}
\end{table*}
\appendix

\section{Label Distribution}
\label{label_dist}
\vspace{-3mm}
We can see the explicit label distribution showing the number of topics belonging to each topic up to level 2 of the category hierarchy tree in Table \ref{table:label_distribution}. 

\vspace{-2mm}
\section{Flat \& Hierarchical Model Architectures}
\label{appendix:archs}
\vspace{-3mm}
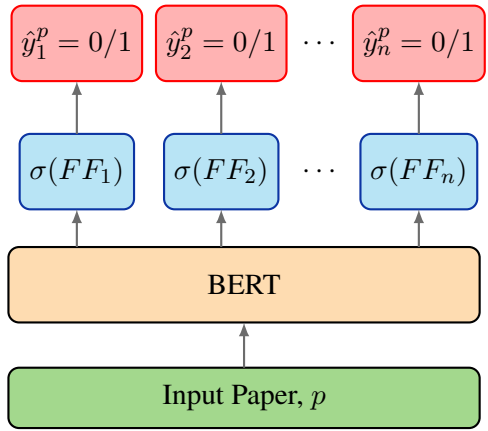
\begin{figure}[t]
\begin{centering}

\vspace{2mm}
%\tikzstyle{entity1}=[thick,text width=15cm] 

\tikzstyle{input_paper}=[shape=rectangle,rounded corners,fill=input,draw=black,thick,minimum width  = 6.2cm,minimum height = .8cm]

\tikzstyle{entity9}=[shape=rectangle,rounded corners,fill=transformersa,draw=black,thick,minimum width  = 6.2cm,minimum height = 1cm] 

\tikzstyle{entity18}=[shape=rectangle,rounded corners,fill=transformerff, draw=mynewblue2,thick,minimum width  = 1.5cm,minimum height = 1cm] 

 \tikzstyle{entity_output}=[shape=rectangle,rounded corners,fill=transformerff, draw=mynewred3,thick,minimum width  = 1.5cm,minimum height = 1cm] 

\begin{tikzpicture} [node distance=6mm][auto,thick][semithick,>=stealth']
 \pgfsetarrowsend{latex}
   
\tikzstyle{entity19}=[shape=rectangle,rounded corners,fill=red!30!white, draw=red,thick,minimum width  = 1.5cm,minimum height = 1cm]

\node[entity9] (z4) at (5.2,3.4) {BERT};
\node[input_paper] (z4) at (5.2,1.9) {Input Paper,  $p$};

\draw[mynewgray,thick] (5.2,2.3) -- (5.2,2.9);

\node[entity18] (z4) at (3.0,4.9) {$\sigma(FF_{1})$};
\node[entity18] (z4) at (4.9,4.9) {$\sigma(FF_{2})$};
\node[entity18] (z4) at (7.5,4.9) {$\sigma(FF_{n})$};
\node (z0) at (6.22, 4.9) {$\cdots$};

\draw[mynewgray,thick] (3.0,3.9) -- (3.0,4.4); 
\draw[mynewgray,thick] (4.9,3.9) -- (4.9,4.4); 
\draw[mynewgray,thick] (7.5,3.9) -- (7.5,4.4);

\node[entity19] (z4) at (3.0,6.6) {$\hat{y}^{p}_{1} = 0/1$};
\node[entity19] (z4) at (4.9,6.6) {$\hat{y}^{p}_{2} = 0/1$};
\node[entity19] (z4) at (7.5,6.6) {$\hat{y}^{p}_{n} = 0/1$};
\node (z0) at (6.22, 6.6) {$\cdots$};

\draw[mynewgray,thick] (3.0,5.4) -- (3.0,6.1); 
\draw[mynewgray,thick] (4.9,5.4) -- (4.9,6.1); 
\draw[mynewgray,thick] (7.5,5.4) -- (7.5,6.1);

%mynewred3

\tikzstyle{entity50}=[shape=rectangle,rounded corners,fill=gray!40!white, draw=black,thick,minimum width  = 7.5cm,minimum height = 0.7cm] 

\end{tikzpicture}

\end{centering}
   \caption{Architecture of our flat multi-label classification baseline using BERT. Here, BERT is the only shared layer for all topics. $FF_{t}$, $\sigma$ and $\hat{y}_{t}^{p}$ denote the feed forward layer for topic $t$, sigmoid activation function and prediction indicating whether topic $t$ is relevant for the input paper $p$ or not (1 or 0), respectively. 
   }
    \label{Flat_Arch}
     %\vspace{-4mm}
\end{figure}

Figure \ref{Flat_Arch} illustrates the architecture of our flat multi-label classification baselines. Here, we show BERT as the encoder to avoid clutter but we also use BiLSTM, XML-CNN and SciBERT as encoders as we describe in Section \ref{sec:methodology}. We can see that the encoder is shared by all topics and there is one feed-forward layer for each topic, $t = 1, 2, ..., n$. A sigmoid activation is applied to the feed-forward layers' output to predict whether each corresponding topic is relevant to an input paper or not (1 or 0). 

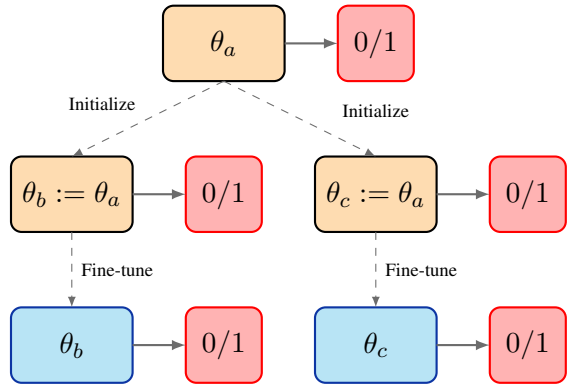
\begin{figure}[t]
\begin{centering}

\vspace{2mm}
%\tikzstyle{entity1}=[thick,text width=15cm] 

\tikzstyle{input_paper}=[shape=rectangle,rounded corners,fill=input,draw=black,thick,minimum width  = 6.2cm,minimum height = .8cm]

\tikzstyle{child_style}=[shape=rectangle,rounded corners,fill=transformerff, draw=mynewblue2,thick,minimum width  = 1.6cm,minimum height = 1cm] 

\tikzstyle{parent_style}=[shape=rectangle,rounded corners,fill=transformersa,draw=black,thick,minimum width  = 1.6cm,minimum height = 1cm]

\tikzstyle{entity_output}=[shape=rectangle,rounded corners,fill=red!30!white, draw=red,thick,minimum width  = 1cm,minimum height = 1cm]

\tikzstyle{entity55}=[shape=rectangle,rounded corners,draw=red,thick,dashed,minimum width  = 7cm,minimum height = 2cm] 

\tikzstyle{entity65}=[shape=rectangle,rounded corners,draw=cyan,thick,dashed,minimum width  = 7cm,minimum height = 2cm]

\begin{tikzpicture} [node distance=6mm][auto,thick][semithick,>=stealth']
 \pgfsetarrowsend{latex}

\node[parent_style] (z4) at (4,5) {$\theta_a$};
\node[entity_output] (z4) at (6.0,5) {$0/1$};
\draw[mynewgray,thick] (4.8,5) -- (5.5,5);

\node[parent_style] (z4) at (2,3) {$\theta_{b} := \theta_{a}$};
\node[entity_output] (z4) at (4,3) {$0/1$};
\draw[mynewgray,thick] (2.8,3) -- (3.5,3);

\node[parent_style] (z4) at (6,3) {$\theta_{c} := \theta_{a}$};
\node[entity_output] (z4) at (8,3) {$0/1$};
\draw[mynewgray,thick] (6.8,3) -- (7.5,3);

%\node (z0) at (6.22, 6.6) {$\cdots$};
\draw[mynewgray,dashed] (4,4.5) -- (2, 3.5);
\draw[mynewgray,dashed] (4,4.5) -- (6, 3.5);

\node[child_style] (z4) at (2,1) {$\theta_b$};
\node[entity_output] (z4) at (4,1) {$0/1$};
\draw[mynewgray,thick] (2.8,1) -- (3.5,1);

\node[child_style] (z4) at (6,1) {$\theta_c$};
\node[entity_output] (z4) at (8,1) {$0/1$};
\draw[mynewgray,thick] (6.8,1) -- (7.5,1);

% \node[entity55] (z4) at (4.7,1) {};
% \node[entity65] (z4) at (4.7,3) {};

\draw[mynewgray,dashed] (2,2.5) -- (2, 1.5);
\draw[mynewgray,dashed] (6,2.5) -- (6, 1.5);

\node (z30) at (2.4,4.2) {\Fontvim{Initialize}};
\node (z30) at (6,4.1) {\Fontvim{Initialize}};

\node (z30) at (2.6,2) {\Fontvim{Fine-tune}};
\node (z30) at (6.6,2) {\Fontvim{Fine-tune}};

\end{tikzpicture}

\end{centering}
   \caption{Leveraging topic hierarchy to learn inter-label dependencies. Here, $\theta_a$, $\theta_b$ and $\theta_c$ are the model parameters of a parent class $a$ and its children $b$ and $c$, respectively. The child models $\theta_b$ and $\theta_c$ are initialized with the parent model $\theta_a$ and then fine-tuned on the data at child nodes $b$ and $c$, respectively. 
   }
    \label{HR_Arch}
     %\vspace{-4mm}
\end{figure}

We can also see an example of leveraging topic hierarchy to learn inter-label dependencies in Figure \ref{HR_Arch}. Here, all models are binary classifiers for a single label from the topic hierarchy. $\theta_a$, $\theta_b$ and $\theta_c$ represent the model parameters of topics $a$, $b$ and $c$, respectively where $a$ is the parent topic of $b$ and $c$ in the hierarchy tree. Both $\theta_b$ and $\theta_c$ are initialized with $\theta_a$ to encode inter-label dependencies and then fine-tuned to predict whether topic $b$ and topic $c$ are relevant to an input paper or not. 

\section{CNN Models and Results}
\vspace{-3mm}
\label{cnn_model_results}
We follow the XML-CNN architecture proposed by \citet{liu2017deep}, which consists of three convolution filters on the word embeddings of the input text. The outputs from the convolution filters are pooled with a certain window. The pooled output then goes trough a bottleneck layer where the dimensionality of the output is reduced to make it computationally efficient. The output from the bottleneck layer is then sent to the output layer for topic prediction. 

 Note that Bi-LSTM, BERT and SciBERT give a contextual representation for every word in the input text which can be used for sequence labeling. This is not necessarily true for CNN. To ensure we have a representation of every word in the input text from CNN filters, the filter sizes are selected in such a way that the number of output vectors match the length of the input text, as presented in  %This idea is adopted from
\cite{xu-etal-2018-double}. Having a corresponding representation for each token is necessary for our multi-task objective. We can see the results of this model in Table \ref{table:cnn_results}.

\vspace{-3mm}
\section{Implementation Details}
\vspace{-3mm}
\label{experimental_Settings}
We started pre-processing our data by converting title, abstract and keywords to lower case letters. Then, we removed the punctuation marks for the LSTM and CNN models. The text was tokenized using the NLTK tokenizer.\footnote{\url{https://www.nltk.org/api/nltk.tokenize.html}} After tokenizing the text, we stemmed the tokens using Porter stemmer.\footnote{\url{https://www.nltk.org/howto/stem.html}} Finally, we masked the words which occur less than two times in the training set with an $<unk>$ tag. The rest of the unique words were used as our vocabulary.

We address the imbalance of classes in our data by assigning the following weights to the examples of the positive class while training our CNN and LSTM based hierarchical classifiers: 1,3,5,10,15...40. The best weight was chosen based on the model's F1 score on the validation set. However, for the flat multi-label classifiers, we could not try this method because finding the optimal combination of weights would take exponential time. We also did not try this approach for hierarchical BERT and Sci-BERT because they are already very time consuming and expensive. We tuned the sigmoid thresholds from $0.1 - 0.9$ on the validation data and the thresholds with the highest performance scores were chosen for every class separately. We tune the loss scaling parameters $[\alpha, \beta]$ for our multi-task objective with the following values: $[0.3, 0.7], [0.4, 0.6], [0.5, 0.5],$ $[0.6, 0.4], [0.7, 0.3], [1, 1]$ on the development set and found that the models show the best performance with $[1, 1]$.

For all our experiments, the maximum lengths for input text (title+abstract) sequence and keywords sequence was set to 100 and 15 respectively. We used pre-trained 300 dimensional Glove\footnote{\url{https://nlp.stanford.edu/projects/glove/}} embeddings to represent the words for LSTM and CNN based models. The hidden state size for the bidirectional LSTMs were kept at 72 across all our models. The fully connected layer after the bi-LSTM layer has size 16 for the hierarchical models and 72 for the flat models. We tried to keep them both at size 16. However, the flat LSTM model showed very unstable performance with a hidden layer of size 16. The filter sizes for the XML-CNN was chosen as 3, 5 and 9 and the number of filters for each of the sizes were set at 64. The input text was padded by 1, 2 and 4 units for each of the filter windows, respectively. The pooling window was set at 32 and the bottleneck layer converted the pooled output to a vector of length 512. 

We used binary cross-entropy as our loss functions for both classification and keyword labeling tasks in all our models. Adam optimizer \cite{kingma2014adam} was used to train the models with mini-batch size 128. Except the transformer based models, the initial learning rate for all other models was kept at 0.001. For BERT and Sci-BERT, the learning rate was set to $2e^{-5}$. The hierarchical LSTM and CNN based models were trained for 10 epochs each. We employed early stopping with patience equal to 3. On the other hand, flat-LSTM and flat-XML-CNN models were trained for 50 epochs with patience 10. The flat and hierarchical transformer based models were fine tuned for 5 and 3 epochs, repectively. We ran our experiments on NVIDIA Tesla K80 GPUs. The average training time was 2 days for the hierarchical LSTM models and additional $\sim$24 hours with the multi-task approach. Hierarchical CNN models took $\sim$24 hours to train with additional 4/5 hours more with the multi-task approach. The flat models took less than 1 hour to train for both LSTM and CNN. The flat transformer based models took $\sim$14 hours to train on one GPU. We used 8 of the same NVIDIA Tesla K80 GPUs to train the hierarchical transformer based models. It took $\sim6$ days to train all 83 binary models.

\clearpage
\onecolumn

%\twocolumn
\begin{small}
\begin{longtable}{ l c c c c}
%\centering
%\small
  %\begin{tabular}{ l c c c}
    \toprule
%      &  \multicolumn{2}{c}{\bf SICK} & {\bf SciTail} & SNLI & clNLI\\%\cline{2-4}
 & & \multicolumn{3}{c}{\bf Count}\\%\cline{2-4}
 \cmidrule(lr){3-5}
 {\bf Category} & {\bf Level} & {\bf \textsc{Train}} & {\bf \textsc{Test}} & {\bf \textsc{Dev}}\\
 \midrule
%CCS & 0 & 148939 & 18617 & 18618\\
General and reference & 1 & 2807 & 365 & 346\\
Hardware & 1 & 6889 & 853 & 866\\
Computer systems organization & 1 & 4758 & 630 & 629\\
Networks & 1 & 8853 & 1102 & 1081\\
Software and its engineering & 1 & 19687 & 2418 & 2518\\
Theory of computation & 1 & 6948 & 824 & 858\\
Mathematics of computing & 1 & 3147 & 410 & 415\\
Information systems & 1 & 25445 & 3167 & 3198\\
Security and privacy & 1 & 4640 & 567 & 609\\
Human-centered computing & 1 & 26309 & 3303 & 3242\\
Computing methodologies & 1 & 20434 & 2602 & 2493\\
Applied computing & 1 & 10491 & 1320 & 1247\\
Social and professional topics & 1 & 8520 & 1055 & 1114\\
Document types & 2 & 488 & 57 & 59\\
Cross-computing tools and techniques & 2 & 1823 & 239 & 223\\
%Printed circuit boards & 2 & 0 & 0 & 0\\
Communication hardware, interfaces and storage & 2 & 1016 & 123 & 131\\
Integrated circuits & 2 & 1571 & 197 & 179\\
Very large scale integration design & 2 & 561 & 82 & 73\\
Power and energy & 2 & 11 & 0 & 1\\
Electronic design automation & 2 & 1279 & 161 & 167\\
Hardware validation & 2 & 826 & 107 & 98\\
Hardware test & 2 & 392 & 37 & 48\\
Robustness & 2 & 289 & 31 & 35\\
Emerging technologies & 2 & 420 & 56 & 65\\
Architectures & 2 & 1941 & 251 & 251\\
Embedded and cyber-physical systems & 2 & 1305 & 173 & 180\\
Real-time systems & 2 & 307 & 48 & 45\\
Dependable and fault-tolerant systems and networks & 2 & 361 & 47 & 59\\
Network architectures & 2 & 1179 & 142 & 132\\
Network protocols & 2 & 1938 & 243 & 213\\
Network components & 2 & 140 & 14 & 17\\
Network algorithms & 2 & 16 & 3 & 4\\
Network performance evaluation & 2 & 26 & 5 & 3\\
Network properties & 2 & 337 & 41 & 39\\
Network services & 2 & 873 & 111 & 126\\
Network types & 2 & 2561 & 340 & 326\\
Software organization and properties & 2 & 4964 & 614 & 586\\
Software notations and tools & 2 & 7018 & 901 & 923\\
Software creation and management & 2 & 5601 & 649 & 735\\
Models of computation & 2 & 270 & 32 & 21\\
Formal languages and automata theory & 2 & 186 & 25 & 28\\
Computational complexity and cryptography & 2 & 189 & 14 & 23\\
Logic & 2 & 642 & 72 & 74\\
Design and analysis of algorithms & 2 & 2444 & 304 & 291\\
Randomness, geometry and discrete structures & 2 & 960 & 111 & 131\\
Theory and algorithms for application domains & 2 & 426 & 45 & 48\\
Semantics and reasoning & 2 & 1001 & 129 & 131\\
Discrete mathematics & 2 & 670 & 89 & 82\\
Probability and statistics & 2 & 665 & 93 & 89\\
Mathematical software & 2 & 214 & 23 & 35\\
Information theory & 2 & 576 & 77 & 64\\
Mathematical analysis & 2 & 653 & 77 & 84\\
Continuous mathematics & 2 & 87 & 8 & 14\\
Data management systems & 2 & 2408 & 319 & 310\\
Information storage systems & 2 & 1159 & 149 & 132\\
Information systems applications & 2 & 8139 & 995 & 972\\
World Wide Web & 2 & 1514 & 191 & 201\\
Information retrieval & 2 & 8697 & 1060 & 1124\\
Cryptography & 2 & 435 & 37 & 42\\
Formal methods and theory of security & 2 & 10 & 2 & 3\\
Security services & 2 & 505 & 46 & 65\\
Intrusion/anomaly detection and malware mitigation & 2 & 293 & 43 & 37\\
Security in hardware & 2 & 30 & 9 & 5\\
Systems security & 2 & 1021 & 120 & 115\\
Network security & 2 & 499 & 68 & 74\\
Database and storage security & 2 & 46 & 3 & 4\\
Software and application security & 2 & 78 & 8 & 7\\
Human and societal aspects of security and privacy & 2 & 94 & 11 & 12\\
Human computer interaction (HCI) & 2 & 13714 & 1777 & 1675\\
Interaction design & 2 & 1079 & 107 & 140\\
Collaborative and social computing & 2 & 2543 & 312 & 324\\
Ubiquitous and mobile computing & 2 & 40 & 4 & 6\\
Visualization & 2 & 127 & 15 & 14\\
Accessibility & 2 & 62 & 7 & 10\\
Symbolic and algebraic manipulation & 2 & 478 & 70 & 75\\
Parallel computing methodologies & 2 & 554 & 69 & 75\\
Artificial intelligence & 2 & 8117 & 1002 & 986\\
Machine learning & 2 & 2567 & 312 & 314\\
Modeling and simulation & 2 & 2123 & 273 & 274\\
Computer graphics & 2 & 3903 & 513 & 459\\
Distributed computing methodologies & 2 & 65 & 14 & 10\\
Concurrent computing methodologies & 2 & 319 & 46 & 56\\
Electronic commerce & 2 & 287 & 38 & 42\\
Enterprise computing & 2 & 438 & 49 & 48\\
Physical sciences and engineering & 2 & 979 & 123 & 121\\
Life and medical sciences & 2 & 1419 & 189 & 171\\
Law, social and behavioral sciences & 2 & 1308 & 159 & 145\\
%Computer forensics & 2 & 0 & 0 & 0\\
Arts and humanities & 2 & 943 & 113 & 106\\
Computers in other domains & 2 & 1219 & 161 & 139\\
Operations research & 2 & 264 & 38 & 28\\
Education & 2 & 1495 & 191 & 180\\
Document management and text processing & 2 & 795 & 97 & 95\\
Professional topics & 2 & 5898 & 736 & 775\\
Computing / technology policy & 2 & 1114 & 161 & 153\\
User characteristics & 2 & 316 & 34 & 37\\

\bottomrule
 \caption{Category name vs paper count up to level 2 of the CCS hierarchy tree in SciHTC.}\\
  %\end{tabular}
 
  %The Macro F1 (\%) of different approaches on our selected datasets. 
    \label{table:label_distribution}
\end{longtable}
\end{small}

\end{document}